\documentclass[conference]{IEEEtran}
\usepackage{cite}
\usepackage{amsmath,amssymb,amsfonts}
\usepackage{algorithmic}
\usepackage{graphicx}
\usepackage{textcomp}
\usepackage{xcolor}
\usepackage{hyperref}
\usepackage{mathrsfs}
\usepackage{amsthm}
\usepackage{booktabs,lipsum}
\usepackage{bbm}
\usepackage[caption=false]{subfig}

\IEEEoverridecommandlockouts

\makeatletter
\newcommand*\titleheader[1]{\gdef\@titleheader{#1}}
\AtBeginDocument{%
  \let\st@red@title\@title
  \def\@title{%
    \bgroup\normalfont\large\centering\@titleheader\par\egroup
    \vskip1.5em\st@red@title}
}
\makeatother
	\title{Edge Entropy as an Indicator of the Effectiveness of GNNs over CNNs for Node Classification \thanks{This material is based upon work partially funded and supported by the Department of Defense under Contract No. FA8702-15-D-0002 with Carnegie Mellon University for the operation of the Software Engineering Institute, a federally funded research and development center; DM20-0590. This work is also partially supported by NSF grant CPS~1837607. The authors are at Carnegie Mellon University \texttt{\{yaoj, jshi3,  markcheu, moura\}@andrew.cmu.edu}, \texttt{owright@sei.cmu.edu}.}} 
	\titleheader{Preprint for 2021 Asilomar Conference}
\begin{document}
	\author{
	\IEEEauthorblockN{Lavender Y. Jiang,   John Shi, Mark Cheung, Oren Wright
 and Jos\'e M.F. Moura}
 \IEEEauthorblockA{Carnegie Mellon University, Pittsburgh, USA }
 }

	\maketitle
	
	\begin{abstract}

 Graph neural networks (GNNs) extend convolutional neural networks (CNNs) to graph based data. A question that arises is how much performance improvement does the underlying graph structure in the GNN provide over the CNN (that ignores this graph structure). To address this question, we introduce edge entropy and evaluate how good an indicator it is for possible performance improvement of GNNs over CNNs. Our results on node classification with synthetic and real datasets show that lower values of edge entropy predict larger expected performance gains of GNNs over CNNs, and, conversely, higher edge entropy leads to expected smaller improvement gains.

\end{abstract}
	\begin{IEEEkeywords}
        Node Classification, Graph Convolutional Neural Network, Interpretability, Geometric Deep Learning 
    \end{IEEEkeywords}
	
	\section{Introduction}
\begin{figure}[hbp!]%
    \centering
    \includegraphics[width=\linewidth]{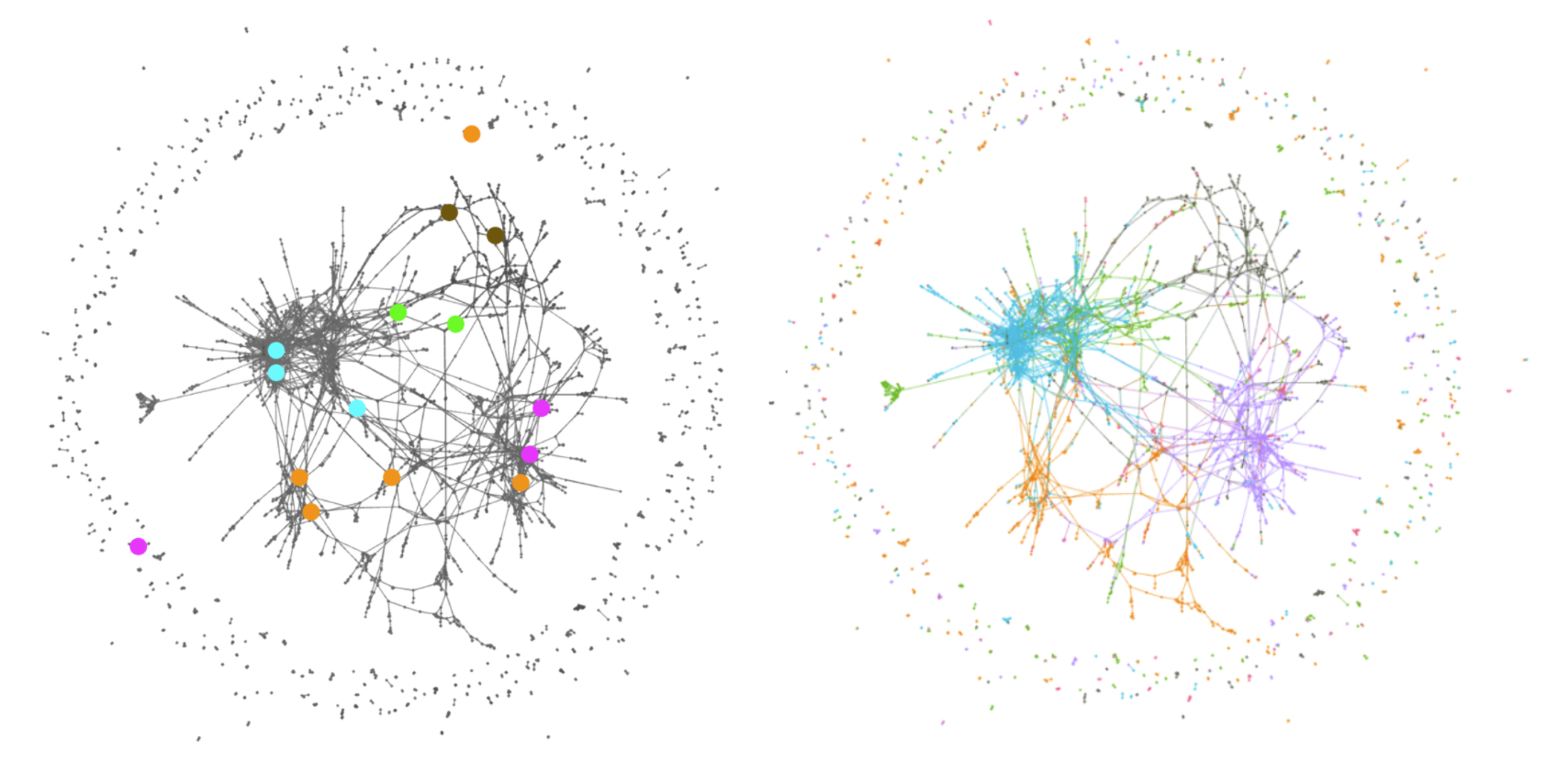}
    \caption{Visualization \cite{GEPHI} of Citeseer \cite{node_dataset} Dataset.}
    \label{figure:citeseer}
    \end{figure}

	Graph Neural Networks (GNNs) extends Convolutional Neural Networks (CNNs) to graph-based data. A common problem solved by GNNs is node classification. Given a partially labelled graph structure and the data defined on each node, the goal of node classification is to accurately classify the unlabelled nodes. For example, Fig.~\ref{figure:citeseer} shows the graph representation of a node classification dataset before and after the node classification, where the colors represent labels of the nodes. On the left, we have a partially labelled graph. On the right, the graph is fully labelled from node classification. The graph structure represents a citation network, where each node represents a paper, and each undirected edge between two nodes represents a citation between the two papers.
    
    While GNNs are commonly used for learning irregular data residing on non-Euclidean domain, it is challenging to evaluate how much the underlying graph structure helps with the model's performance. Node classification problems can be solved using either a GNN or a CNN\footnote{In this case, the CNN is similar to a Multilayer Perceptron (MLP) because the sliding window of convolution layers has size 1.}. One benefit of using a GNN is that it uses the given graph structure to classify nodes, while a CNN does not. However, we do not know whether using the graph structure and a GNN will provide any advantages over using a CNN and no graph structure a priori. The graph structure could provide little to no information about the labels. This motivates our question: How can we evaluate the effectiveness of the graph structure for GNNs?

\begin{figure}[h!]
	\centering
		\includegraphics[width=7cm]{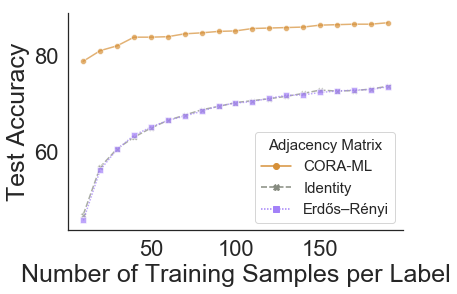}
		\caption{Comparison of Test Accuracy of CORA-ML Dataset \cite{node_dataset} of a 2-layer TAGCN\cite{TAGCN} with 1) given dataset's graph, 2) identity matrix, 3) random Erd\H{o}s-R\`enyi graph \cite{SPM}.}
		\label{figure:compare}
	\end{figure}

    To answer this question, we need to address two aspects. First, we need to define what makes a graph structure effective. We say a graph structure is relatively effective if it contains a relatively large amount of information about the label distribution. This information added by the graph structure is reflected in test accuracies. For instance, Fig.~\ref{figure:compare} shows the test accuracies of a GNN model on a citation network dataset using three different graph structures. We say the underlying graph structure of CORA-ML is more effective than the random graph, because while the edges in CORA-ML represent citations, the edges in the random graph are randomly drawn and convey no useful information. Similarly, the test accuracy of the model trained with the underlying graph structure (orange solid line) is around 20 percent higher than the model trained with the identity matrix (grey dashed line), whereas the test accuracy of the model trained with the random graph (purple dotted line) has almost no improvement from the model trained with the identity matrix.

	 The second question we need to address is, how do we quantitatively evaluate the effectiveness of graph structure? In this paper, we propose an information theoretic parameter called edge entropy to measure the quality of label information contained in the graph structure. We show edge entropy is a good indicator of the effectivenss of GNN over CNN using experiment results from both synthetic and real datasets.

     \section{Related Work}
     Node classification has been used in many different settings, but very little has been done on evaluating the effectiveness of graph structure in the node classification datasets. Much of the existing work on evaluating GNN models focuses on developing benchmarks and upper bounding the modeling capacity for graph classification tasks. Dwivedi, Joshi, Laurent, Bengio and Bresson \cite{dwivedi2020benchmarkgnns} introduced a standardized benchmarking framework for running a variety of GNN experiments. Xu, Hu, Leskovec and Jegelka \cite{powerful} showed that a Graph CNN is only as powerful as the Weisfeiler-Lehman graph isomorphism test for graph classification tasks. 
     
     Another related but different problem is graph comparison. Metrics such as the clustering coefficient \cite{SmallWorld}, the betweenness centrality \cite{centrality}, and graph spectral distance \cite{spectral} were proposed to discern the topological properties of different graph structures. This problem is different than ours, because these metrics do not consider the nodes' labels. Our challenge is to evaluate the effectiveness of the given graph structure based on both the connectivity and the label distribution.

    \section{Problem Statement}
    
   We are interested in evaluating the effectiveness of GNN over CNN by examining the effectiveness of the graph structure in node classification datasets. In this paper, we focus on a variant of GNN called Graph Convolution Neural Networks (Graph CNNs), where the convolution layer is explicitly defined using the adjacency matrix. 
   
   \subsection{Node Classification}
   
   A node classification dataset $D$ contains the following:
   \begin{itemize}
       \item the graph structure, i.e., the adjacency matrix $A$
       \item the data defined on each node $X$
       \item the labels for all nodes $y$\textbf{}
   \end{itemize}
The goal of node classification is to accurately classify unlabelled nodes based on $A,X$, and part of $y$.
   
   \subsection{Filter-based Graph CNN}
   We focus on filter-based graph CNN models such as TAGCN\cite{TAGCN} and GCN\cite{GCN}.  Filter-based graph CNN models represent convolution as a filtering operation with a polynomial of the graph shift $S$, where $S$ can be either the adjacency matrix $A$ or the graph Laplacian $L$ \cite{Overview}. Formally, an adjacency matrix based graph convolution layer learns a graph filter \cite{TAGCN}:
    \begin{equation}
    P(A)(X) = \sum_{k=0}^{d-1} A^{k} X W_{k}
    \label{eq:conv}
    \end{equation}
    where $P(A)$ is a polynomial of $A$, $X$ is the input graph signal, $d$ is the degree of the graph polynomial, and $\{W_{k}\}_{k=0}^{d-1}$ are learnable weights. 
    
    \subsection{Effectiveness of Graph Structures}
    For a node classification problem, a graph structure is less effective if the graph structure contains little information about the label distribution. This lack of information can be seen in the test accuracies of the graph CNN model. For a less effective graph structure, the model trained with the given adjacency matrix will yield little improvement in test accuracy compared with the model trained with the identity matrix, i.e., trained with the features only. 
    For example, suppose we have two node classification datasets $D_{1}$ and $D_{2}$. We train a graph CNN model on both datasets using the given adjacency matrix. We also train the model for both datasets using the identity matrix instead of the adjacency matrix.
    
    We define the improvement on $D_i$ as the difference in test accuracy when using the given adjacency matrix versus using the identity matrix. 
    If the improvement on $D_1$ is greater than the improvement on $D_2$, then the graph structure in $D_1$ provides more information, i.e., is more effective, than the graph structure in $D_2$. Likewise, if the improvement on $D_2$ is greater, then the graph structure in $D_2$ is more effective than the graph structure in $D_1$.
    
    \section{Proposed Method}
    To evaluate the effectiveness of the graph structure in a node classification task, we propose a parameter to evaluate the quality of label information contained in the graph structure. We will first motivate our definition with some simple examples, then we will define the parameter.
    
    \subsection{Motivating Examples}
    \label{sec:examples}

\begin{figure}[h!]
    	\centering
		\includegraphics[width=8cm]{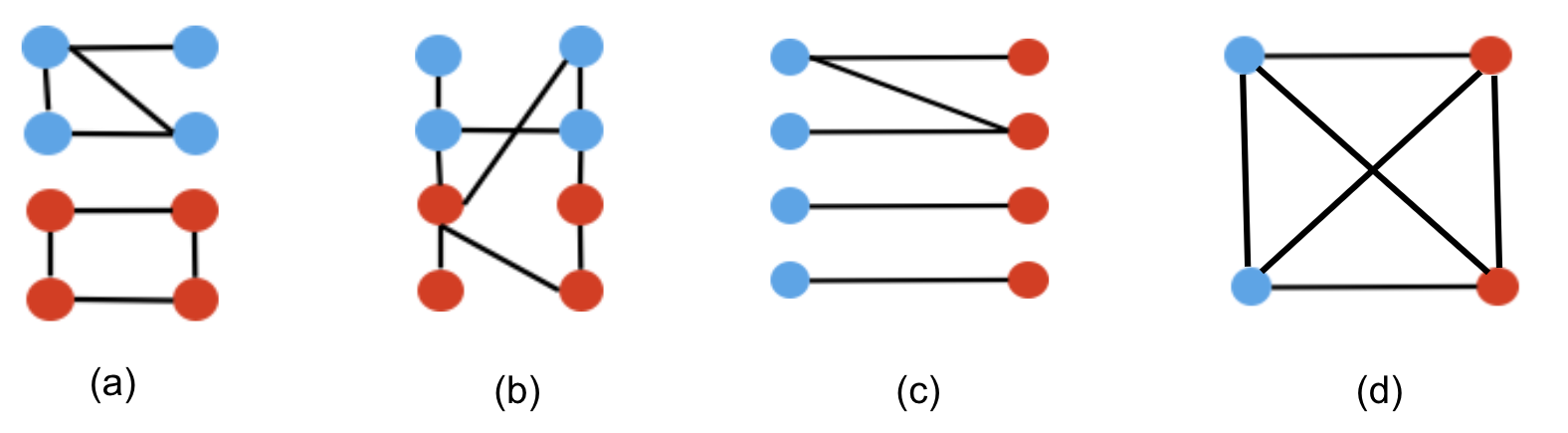}
		\caption{Simple graphs.}
		\label{figure:graphs}
	\end{figure}

    A naive attempt at evaluating the effectiveness of a graph structure is related to node clustering. Consider the graph in Fig. \ref{figure:graphs}(a). There is a blue cluster and a red cluster. Here, blue nodes always connect with blue nodes and red nodes always connect with red nodes. In this case, the edges reveal a lot of label information. For instance, suppose we want to classify an unlabelled node. If we know it has a blue neighbor, then we can confidently classify this node as blue, because the blue cluster and the red cluster are disjoint. 
    
    On the other hand, consider the graph in Fig. \ref{figure:graphs}(b). Here, the blue and red clusters are mixed. It is equally likely for a blue node to connect with either another blue node or a red node. In this case, the edges do not contain useful label information. For instance, knowing that an unlabelled node has a blue neighbor does not make that node more likely to be blue or red. 
    
	A simple parameter inspired by this observation is intra-class ratio, which measures the percentage of edges that connect nodes of the same class. One might hypothesize that a graph with high intra-class ratio is easier to classify with the underlying graph structure, because the clusters are more separated. However, a counter-example for this hypothesis is shown in Fig. \ref{figure:graphs}(c). In this bipartite graph, blue nodes always connect with red nodes and red nodes always connect with blue nodes. In this case, the intra-class ratio is zero, but the graph is very helpful for node classification. For instance, if an unlabelled node has a blue neighbor, then we know it must be a red node.  
	
	Another parameter to consider is the clustering coefficient, which measures how likely nodes would form cliques in the graph. One might assume that a higher clustering coefficient indicates that accounting for the graphs will be more helpful for node classification, because the clusters are more obvious. However, Fig.~\ref{figure:graphs}(d) shows a counter-example for this observation. In this 4-clique graph, the clustering coefficient is 1, but the graph structure does not reveal label information. This is because we have a mixed cluster of both blue and red nodes. In this case, the clustering coefficient provides little help with evaluating the effectiveness of graph structure for node classification, because it ignores the labels of nodes. 
	
	To account for both the connections between different classes and the labels of nodes, we propose a more comprehensive parameter called edge entropy. 
    \subsection{Edge Entropy}
    
    We propose edge entropy as a parameter to evaluate the impact of accounting for the graph structures in node classification. Edge entropy measures the quality of label information encoded in the graph. 
    A high edge entropy (closer to 1) indicates that accounting for the graph structure is very helpful for node classification. A low edge entropy (closer to 0) indicates that accounting for the graph structure is not very useful for node classification.

    Formally, given a graph $\mathcal{G}$ with~$M$ classes of nodes, we define the \textbf{per-class edge entropy} of any class $l$ as a function $H: \{1, \hdots, M\} \to [0,1]$ such that 
\begin{equation}\label{edgeEntropy}
	H(l) := -\sum_{m \in \{1, \hdots, M\}} p_{lm}(n)\log_{M}(p_{lm}(n))
\end{equation}
where the first order interclass connectivity probability $p_{lm}$ is defined as
\begin{equation}\label{connProb}
p_{lm} := \frac{|\{\text{edge } w: \text{start}(w) \in \mathcal{V}_{l} \land \text{end}(w) \in \mathcal{V}_{m} \}|}{|\{\text{edge } w: \text{start}(w) \in \mathcal{V}_{l}\}|}
\end{equation}
where an edge is a member of $\mathcal{E}$ and not a self loop, $\mathcal{V}_{l}$ is the set of nodes that belong to the $l$-th class. Specifically, $p_{lm}(n)$ is the probability that a node~$v$ belongs to class~$l$ given that $v$ is a direct neighbor of a node of class~$l$. 

For a dataset, we define its \textbf{edge entropy} as:
\begin{equation}\label{edgeEntropyFull}
	\widehat{H} := \sum_{m \in \{1, \hdots, M\}} H(m)w_{m}
\end{equation}
where $w_{m}$ is the percentage of samples from class $m$. That is, the ratio between the number of class $m$ nodes and the number of total nodes.

For instance, we can analyze the edge entropies of the graphs in our motivating examples. The two graphs in Fig. \ref{figure:graphs}(a) and  Fig. \ref{figure:graphs}(c) have $\widehat{H}=0$. The graph in Fig. \ref{figure:graphs}(b) and Fig. \ref{figure:graphs}(d) have $\widehat{H}=1$, because the edges are random. This shows that edge entropy is a good indicator of the effectiveness of graph structures for simple graph structures.

\section{Experimental Evaluation}
In this section, we show the correlation between edge entropy and accuracy gains of GNNs from accounting the underlying graph structures using both synthetic and real datasets. We also compare edge entropy with other parameters such as the clustering coefficient and the intra-class ratio. We will start by explaining the datasets and experimental setup, then we will discuss the results from graph CNN experiments. 

\subsection{Synthetic Datasets}

We generate synthetic datasets with specific edge entropies. To accomplish this, we use the following approach. 

Let $N$ be the number of nodes. Let $M$ be the number of classes. Let $r_i$, $i = 1,\hdots,M$ be the number of nodes with class $i$ where $\sum_{i=1}^{M}r_{i}=N$. Let $\rho$ be a sparsity factor, $0\leq \rho \leq 1$.

In order to have a specific edge entropy, the generated connected graph has to have a specific number of edges between nodes of each class. We create a $M \times M$ matrix $T$ with the desired edge entropy where $T_{i,j}$ is the number of edges that connect a class $i$ node to a class $j$ node. We normalize each row of $T$ to produce a probability matrix $P$. $P$ is the matrix of $p_{lm}$ defined for each pair of classes in \eqref{connProb}. Using the values in $P$ as $p_{lm}$ in \eqref{edgeEntropy} and \eqref{edgeEntropyFull} yields the desired edge entropy $\widehat{H}$.

For example, to get an edge entropy of approximately $\widehat{H} = 0.24$ with $M = 2$ classes with $N = 100$ nodes: $r_1 = 50$ class 1 nodes and $r_2 = 50$ class 2 nodes, a possible $T$ and $P$ are $T = \begin{bmatrix} 48 & 2 \\ 2 & 48 \end{bmatrix}$, $P = \begin{bmatrix} 0.96 & 0.04 \\ 0.04 & 0.96 \end{bmatrix}$\footnote{From \eqref{edgeEntropy} and \eqref{edgeEntropyFull}, ${H(0) = H(1) = -(0.96 \log_2{0.96} + 0.04\log_2{0.04})}$ $ = 0.24$ and $\widehat{H}=0.24 \times 0.5+0.24 \times 0.5=0.24$}.

To create the graph, we start with $N$ isolated nodes. Each node has a self loop and random features. A label is assigned to each node, based on the distribution of nodes by class, $r_i$, $i = 1,\hdots,M$.
We then consider every pair of nodes $v_i$, $v_j$, $i,j = 1,\hdots,N$. For each pair of nodes, we create a directed edge from $v_i$ to $v_j$ with probability $\rho P_{l,m}$ where $l$ is the class of $v_i$ and $m$ is the class of $v_j$.

\subsection{Other Datasets} 

We used a variety of popular node classfication datasets :

\begin{itemize}
    \item Citation network datasets including CORA-ML, CITESEER and PUBMED \cite{node_dataset}. Nodes are scientific papers and edges represent citations between papers. Data defined on each node is a bag of words vector. Labels represent the field of study. 
    \item Social network datasets such as REDDIT \cite{reddit}. Nodes represent users' posts and an edge is drawn between two posts if the same user commented on both of the posts. There is no data defined on each node. Labels represent which subreddit a post belongs to. 
    \item Large synthetic datasets such as SBM-PATTERN and SBM-CLUSTER. Each dataset consists of 10,000 randomly generated graphs using Stochastic Block Models. They were introduced as a benchmark dataset for evaluating GNNs in \cite{dwivedi2020benchmarkgnns}.
\end{itemize}

\subsection{Experimental Setup}
For each  dataset, we trained a GNN model using both the underlying graph structure and the identity matrix. We then compare the difference between the test accuracy when we train with the underlying graph structure and the test accuracy when we train with the identity matrix. We call this difference in accuracy the \textbf{improvement} on the dataset from accounting for the graph structure. 

\subsubsection{Synthetic Datasets} We generate the graphs using the following parameters: $N=3000, M=3, r_{1}=r_{2}=r_{3}=1/3$. For comparison purposes, we choose two sparsity factor $\rho_{1}=0.1, \rho_{2}=0.5$. The synthetic graph generated with $p_{1}$ is sparse, and a graph generated with $p_{2}$ is dense. We also fix two connection probability matrices such that one wo;; have low edge entropy ($\widehat{H}_{\text{low}}\approx 0.52$), and the other one will have high edge entropy ($\widehat{H}_{\text{high}}\approx 0.97$). Specifically, 
\[P_{\text{low}} = \begin{bmatrix}
0.8 & 0.05 & 0.15\\
0.05 & 0.9 & 0.05 \\
0.27 & 0.03 & 0.7 
\end{bmatrix}, \,
P_{\text{high}} = \begin{bmatrix}
0.4 & 0.26 & 0.34\\
0.2 & 0.5 & 0.3 \\
0.33 & 0.31 & 0.37 
\end{bmatrix}
\]

Dataset dense\_low is generated with $\rho_{2}, P_{\text{low}}$. Dataset sparse\_low is generated with $\rho_{1}, P_{\text{low}}$. Dataset dense\_high is generated with $\rho_{2}, P_{\text{high}}$. Dataset sparse\_high is generated with $\rho_{1}, P_{\text{high}}$. For each dataset, we run 100 Monte Carlos trials by training a randomly initialized 2-layer TAGCN with 2nd order filters for 200 epochs, and testing with cross validation. We vary our percentage of training data from 10\% to 90\% with a 10\% increment each step.

\subsubsection{Other Datasets}
For citation network datasets, we trained a 2-layer TAGCN with polynomial order 3 and 16 hidden units using cross validation. We used an Adam optimizer with a learning rate of 0.01 and a learning rate decay factor of 5e-4. 

For the Reddit dataset, we trained a 2-layer Graph Attention Network (GAT) \cite{Velickovic2018}, because we need the sampling function to handle the large size of the dataset. We used 8 heads in the first layer and 1 head in the second layer, a dropout rate of 0.6 for both layers,  200 epochs, and an Adam optimizer with a learning rate of 5e-3 and a learning rate decay factor of 5e-4. For comparison purposes we also trained the same model on citation network datasets.

For the SBM datasets, we trained a 1-layer GCN with 146 hidden units and 1000 learning epochs. We used an Adam optimizer with a learning rate of 1e-3 and a learning rate reduce factor of 0.5.

\begin{figure}[h!]
\centering

\subfloat[The upper blue line represents the test accuracy on dataset sparse\_low. The lower orange line represents the test accuracy on dataset sparse\_high.]{%
  \includegraphics[clip,width=0.7\columnwidth]{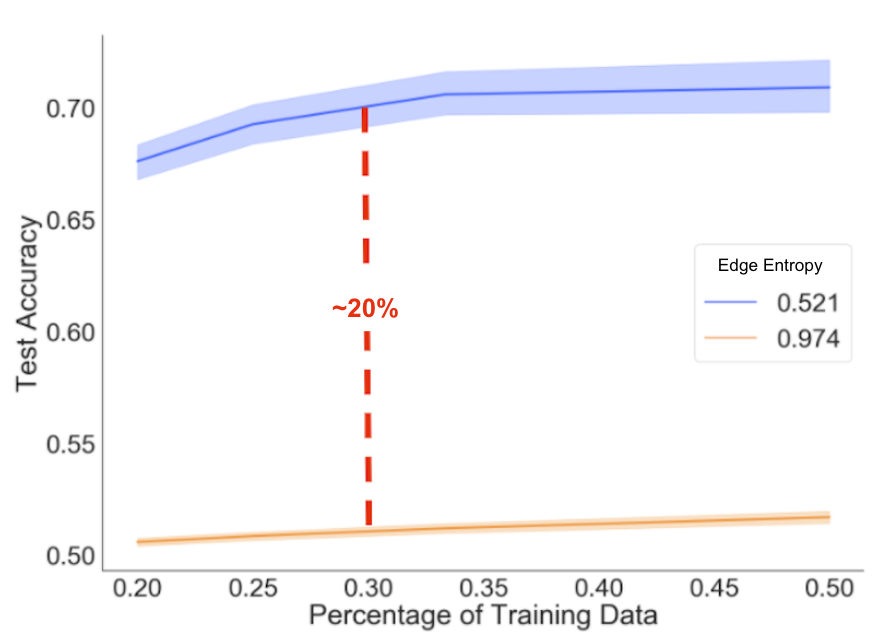}%
}

\subfloat[The upper blue line represents the test accuracy on dataset dense\_low. The lower orange line represents the test accuracy on dataset dense\_high]{%
  \includegraphics[clip,width=0.7\columnwidth]{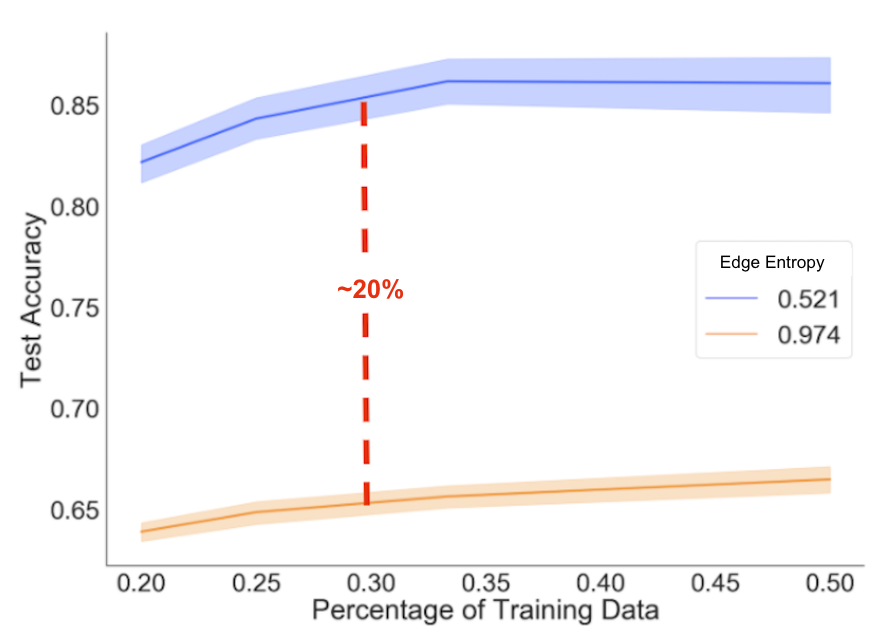}%
}
		\caption{Test accuracy of TAGCN on synthetic datasets. The red dashed line shows the improvement from accounting the graph structure is around 20\% with 30\% training samples for both sparsity factors. The shaded regions represent standard deviation of the Monte Carlos trial outcomes. This shows that edge entropy is a good indicator of the effectiveness of the graph structure in synthetic datasets.}
		\label{figure:syn_acc}
\end{figure}

	\begin{table}[h!]
    \centering
    \caption{Accuracy Improvements with 2-Layer TAGCN for Synthetic Datasets with 30 Percent Training Data}
    \begin{tabular}{|c|c|c|c|c|c|c|c|}
    \hline
     Dataset &  Clustering & Intra   & $\widehat{H}$ & Improvement\\ \hline
     dense\_low & \textbf{0.45} & \textbf{0.8} & \textbf{0.521}  & \textbf{51.3}  \\ 
     \hline
    sparse\_low & 0.121 & \textbf{0.8} & \textbf{0.521}  & 36.2  \\ 
    \hline
    dense\_high & 0.31 & 0.42 & 0.974 & 32  \\ 
    \hline
    sparse\_high & 0.077 & 0.42 & 0.974 & 19.2  \\ 
    \hline
    \end{tabular}
    \label{table:synth}
    \end{table}




\begin{table}[!htbp]
    \centering
      \caption{Accuracy Improvements with 2-layer TAGCN for Classifying Citation Networks}
    \begin{tabular}{|c|c|c|c|c|c|c|c|}
    \hline
     Dataset & Clustering & Intra & $\widehat{H}$ & Improvement\\ \hline
    CORA-ML & 0.242  & \textbf{0.81} &  $\textbf{0.390}$ & $\textbf{63.9}$\\ 
     \hline
    CiteSeer & 0.144 & 0.74 &  0.533  & 54.9  \\ 
    \hline
    Pubmed & 0.066 & 0.80 & 0.564 & 51 \\ 
    \hline
    \end{tabular}
    \label{table:citations}
    \end{table}

	\begin{table}[!htbp]
    \centering
    \caption{Accuracy Improvements with 2-Layer GAT for Social Networks and Citation Networks}
    \begin{tabular}{|c|c|c|c|c|c|c|c|}
    \hline
     Dataset &  Clustering & Intra   & $\widehat{H}$ & Improvement\\ \hline
     Reddit & \textbf{0.32} & 0.756 & $\textbf{0.322}$ & $\textbf{45.1}$  \\ 
    \hline
     CORA-ML & 0.242 & \textbf{0.81} & 0.390  & 23  \\ 
     \hline
    CiteSeer & 0.144 & 0.74 & 0.533  & 13.5  \\ 
    \hline
    PubMed & 0.066 & 0.80 & 0.564 & 7.4  \\ 
    \hline
    \end{tabular}
    \label{table:gat}
    \end{table}

     \begin{table}[!htbp]
    \centering
          \caption{Accuracy Improvements with 1-Layer GCN for SBM Datasets}
    \begin{tabular}{|c|c|c|c|c|c|c|}
    \hline
     Dataset & Clustering &Intra & $\widehat{H}$ & Improvement \\ \hline
    SBM\_PATTERN & \textbf{0.427} & \textbf{0.589} & \textbf{0.811} & \textbf{34.78} \\ 
    \hline
    SBM\_CLUSTER & 0.317 &0.33 & 0.954 & 27.68 \\ 
     \hline
    \end{tabular}
    \label{table:sbm}
    \end{table}
    
    \subsection{Discussions}
    The results with both synthetic and real datasets show edge entropy is a good indicator of the effectiveness of accounting the graph structure in node classification. For example, Fig.~\ref{figure:syn_acc} shows that the synthetic dataset with low edge entropy always have higher improvement than the synthetic dataset with high edge entropy for both sparsity factors, $\rho_1 = 0.1, \rho_2 = 0.5$. Datasets dense\_low (Table~\ref{table:synth}), CORA-ML (Table~\ref{table:citations}), Reddit (Table~\ref{table:gat}) and SBM\_PATTERN (Table~\ref{table:sbm}) also have the lowest edge entropy and the highest improvement in their tables. This shows that a lower edge entropy corresponds to a higher improvement in test accuracy, and a higher edge entropy corresponds to a lower improvement in test accuracy.
    
    Other parameters such as the clustering coefficient and the intra-class ratio are not consistent with the impact of accounting for graph structures in certain cases. For instance, the clustering coefficient for dataset sparse\_low in Table \ref{table:synth} is 0.121. This is lower than the clustering coefficient for dense\_high (0.31) and does not reflect the better improvement of sparse\_low (51.3\%) over dense\_high (32\%). Another example is the intra-class ratio for Reddit in Table \ref{table:gat}. The intra-class ratio for Reddit is 0.756, which is higher than the intra-class ratio of CORA-ML (0.242). This is inconsistent with Reddit's higher improvement (45.1\%) from accounting the graph structure than CORA-ML (23\%). 
    The underlying reason is that the clustering coefficient ignores node labels and the intra-class ratio neglects connections between different classes, as we discussed in the motivating examples. 
    
    Another interesting observation is that while the clustering coefficient is inconsistent with the improvements on synthetic datasets, a high clustering coefficient indicates a higher improvement for the real datasets studied in this paper. A possible explanation is that these problems are similar to node clustering. This might help explain why a method as simple as label propagation \cite{labelprop} can achieve state-of-the-art accuracy on some popular node classification datasets.

\section{Conclusion}
Evaluating the effectiveness of GNN over CNN on node classification datasets is an important issue, because it shows advantages of using a graph structure. In this paper, we defined what makes the graph structure effective for node classification, and we proposed edge entropy as a parameter to quantitatively evaluate the effectiveness of graph structures. We showed edge entropy is a good indicator of the effectiveness of the graph with experiment results from both synthetic and real datasets. Alternative parameters such as the clustering coefficient and the intra-class ratio are inconsistent with the accuracy gains from accounting the graph structure in some cases. In future works, we will extend the definition of edge entropy to consider longer walks of lengths greater than 1, and we will improve our simulations with other graph models such as small-world graphs and preferential attachments.

\section{Acknowledgments}
The authors thank Xujin Chris Liu, Srinivasa Pranav, Rajshekhar Das, Chaojing Duan for helpful discussions.

\bibliographystyle{IEEEtran}
\bibliography{refs}

\begin{thebibliography}{10}
\providecommand{\url}[1]{#1}
\csname url@samestyle\endcsname
\providecommand{\newblock}{\relax}
\providecommand{\bibinfo}[2]{#2}
\providecommand{\BIBentrySTDinterwordspacing}{\spaceskip=0pt\relax}
\providecommand{\BIBentryALTinterwordstretchfactor}{4}
\providecommand{\BIBentryALTinterwordspacing}{\spaceskip=\fontdimen2\font plus
\BIBentryALTinterwordstretchfactor\fontdimen3\font minus
  \fontdimen4\font\relax}
\providecommand{\BIBforeignlanguage}[2]{{%
\expandafter\ifx\csname l@#1\endcsname\relax
\typeout{** WARNING: IEEEtran.bst: No hyphenation pattern has been}%
\typeout{** loaded for the language `#1'. Using the pattern for}%
\typeout{** the default language instead.}%
\else
\language=\csname l@#1\endcsname
\fi
#2}}
\providecommand{\BIBdecl}{\relax}
\BIBdecl

\bibitem{GEPHI}
M.~Bastian, S.~Heymann, and M.~Jacomy, ``Gephi: An open source software for
  exploring and manipulating networks,'' in \emph{International AAAI Conference
  on Weblogs and Social Media}, 2009.

\bibitem{node_dataset}
P.~Sen, G.~Namata, M.~Bilgic, L.~Getoor, B.~Galligher, and T.~Eliassi-Rad,
  ``Collective classification in network data,'' \emph{AI Magazine}, vol.~29,
  no.~3, 2008.

\bibitem{TAGCN}
J.~Du, S.~Zhang, G.~Wu, J.~M.~F. Moura, and S.~Kar, ``Topology adaptive graph
  convolutional networks,'' \emph{Computing Research Repository}, vol.
  abs/1710.10370, 2017.

\bibitem{SPM}
M.~Cheung, J.~Shi, O.~Wright, Y.~L. Jiang, X.~Liu, and J.~M.~F. Moura, ``Graph
  signal processing and deep learning: Convolution, pooling, and topology,''
  \emph{IEEE Signal Processing Magazine}, vol.~37, no.~6, pp. 139--149, 2020.

\bibitem{dwivedi2020benchmarkgnns}
V.~P. Dwivedi, C.~K. Joshi, T.~Laurent, Y.~Bengio, and X.~Bresson,
  ``Benchmarking graph neural networks,'' \emph{Computing Research Repository},
  vol. abs/2003.00982, 2020.

\bibitem{powerful}
K.~Xu, W.~Hu, J.~Leskovec, and S.~Jegelka, ``How powerful are graph neural
  networks?'' in \emph{7th International Conference on Learning
  Representations}, 2019.

\bibitem{SmallWorld}
D.~J. Watts and S.~H. Strogatz, ``Collective dynamics of `small-world'
  networks,'' \emph{Nature}, vol. 393, no. 6684, pp. 440--442, Jun. 1998.

\bibitem{centrality}
L.~C. Freeman, ``A set of measures of centrality based on betweenness,''
  \emph{Sociometry}, vol.~40, no.~1, pp. 35--41, 1977.

\bibitem{spectral}
I.~Jovanovi{\'{c}} and Z.~Stani{\'{c}}, ``Spectral distances of graphs,''
  \emph{Linear Algebra and its Applications}, vol. 436, no.~5, pp. 1425--1435,
  Mar. 2012.

\bibitem{GCN}
T.~N. Kipf and M.~Welling, ``Semi-supervised classification with graph
  convolutional networks,'' in \emph{5th International Conference on Learning
  Representations}, 2017.

\bibitem{Overview}
A.~{Ortega}, P.~{Frossard}, J.~{Kovačević}, J.~M.~F. {Moura}, and
  P.~{Vandergheynst}, ``Graph signal processing: Overview, challenges, and
  applications,'' \emph{Proceedings of the IEEE}, vol. 106, no.~5, pp.
  808--828, May 2018.

\bibitem{reddit}
P.~Yanardag and S.~Vishwanathan, ``Deep graph kernels,'' in \emph{Proceedings
  of the 21th ACM SIGKDD International Conference on Knowledge Discovery and
  Data Mining}, ser. KDD '15.\hskip 1em plus 0.5em minus 0.4em\relax
  Association for Computing Machinery, 2015, p. 1365–1374.

\bibitem{Velickovic2018}
P.~Veličković, G.~Cucurull, A.~Casanova, A.~Romero, P.~Li{\`{o}}, and
  Y.~Bengio, ``{Graph attention networks},'' in \emph{6th International
  Conference on Learning Representations (ICLR)}, 2018.

\bibitem{labelprop}
\BIBentryALTinterwordspacing
Anonymous, ``Combining label propagation and simple models out-performs graph
  neural networks,'' in \emph{Submitted to International Conference on Learning
  Representations}, 2021, under review. [Online]. Available:
  \url{https://openreview.net/forum?id=8E1-f3VhX1o}
\BIBentrySTDinterwordspacing

\end{thebibliography}
\end{document}